\pdfoutput=1
\documentclass[11pt,a4paper]{article}
\usepackage[hyperref]{acl2020}
\usepackage{times}
\usepackage{latexsym}

\usepackage{graphicx}
\usepackage{url}
\usepackage{subcaption}
\usepackage{amsmath,bm,amsfonts,dsfont,bbm}
\usepackage{paralist}
\usepackage{hyperref}
\usepackage{multicol}
\usepackage{color}
\usepackage{booktabs}
\usepackage[inline]{enumitem}
\usepackage{xspace}
\usepackage{mdframed}

\usepackage[activate={true,nocompatibility}, kerning=true,spacing=true, stretch=30,shrink=30]{microtype}
\microtypecontext{spacing=nonfrench}
\aclfinalcopy %

\newcommand{\actignore}{\textsc{ignore}\xspace}
\newcommand{\actcoref}{\textsc{coref}\xspace}
\newcommand{\actoverwrite}{\textsc{overwrite}\xspace}
\newcommand{\mysim}{\mathit{sim}}
\newcommand{\mlp}{\mathrm{MLP}}
\newcommand{\cs}{\mathit{cs}}

\newcommand{\bertbase}{BERT\textsubscript{BASE}\xspace}
\newcommand{\bertlarge}{BERT\textsubscript{LARGE}\xspace}
\newcommand{\modelname}{PeTra\xspace}
\newcommand\Mark[1]{\textsuperscript#1}
\newcommand{\hlent}[2]{\colorbox{gray!30}{#1\textsubscript{#2}}}

\definecolor{dkgreen}{RGB}{0,130,0}
\definecolor{aqua}{rgb}{0.0, 0.4, 1.0}

\def\vec#1{\ensuremath{\boldsymbol{{#1}}}}

\title{PeTra: A Sparsely Supervised Memory Model for People Tracking}%

\author{Shubham Toshniwal\Mark{1},
Allyson Ettinger\Mark{2},
Kevin Gimpel\Mark{1},
Karen Livescu\Mark{1}\\
\Mark{1}Toyota Technological Institute at Chicago\\
\Mark{2}Department of Linguistics, University of Chicago\\[0.5em]
\small{\texttt{\{shtoshni, kgimpel, klivescu\}@ttic.edu, aettinger@uchicago.edu}}\\
}

\date{}

\begin{document}
\maketitle

\begin{abstract}
We propose \modelname, a memory-augmented neural network designed to track entities in its memory slots. %
\modelname is trained using sparse annotation from the GAP pronoun resolution dataset and outperforms a prior memory model on the task while using a simpler architecture.
We empirically compare key modeling choices, finding that we can simplify several aspects of the design of the memory module
while retaining strong performance.
To measure the people tracking capability of memory models, we (a) propose
a new diagnostic evaluation based on counting the number of unique entities in text,
and (b) conduct a small scale human evaluation to compare evidence of people tracking in the memory logs of \modelname relative to a previous approach. %
\modelname is highly effective in both evaluations, demonstrating its ability to track people in its memory despite being trained with  %
limited annotation.
\end{abstract}

\begin{figure*}[ht]
    \centering
     \includegraphics[width=\textwidth]{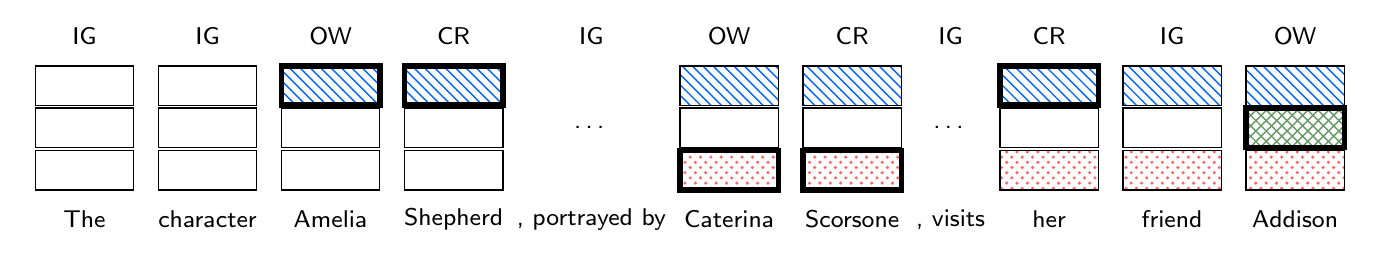}
     \captionof{figure}{Illustration of memory cell updates in an example sentence where IG = ignore, OW = overwrite, CR = coref. Different patterns indicate the different entities, and an empty pattern indicates that the cell has not been used. The updated memory cells at each time step are highlighted.
     }
     \label{fig:ideal_run}
\end{figure*}

\section{Introduction}
Understanding text narratives requires maintaining and resolving entity references over arbitrary-length spans.
Current approaches for coreference resolution  \cite{clark-manning-2016-improving,lee-etal-2017-end, lee-etal-2018-higher, wu2019coreference} scale quadratically (without heuristics) with length of text, and hence are impractical for long narratives.
These models are also cognitively implausible, lacking the incrementality of human language processing~\cite{Tanenhaus1632, keller-2010-cognitively}.
Memory models with finite memory and online/quasi-online entity resolution have linear runtime complexity, offering more scalability, cognitive plausibility, and interpretability.

Memory models
can be viewed as general problem solvers with external memory mimicking a Turing tape~\cite{graves2014neural,graves2016hybrid}.
Some of the earliest applications of memory networks in language understanding were
for question answering, where the external memory simply stored all of the word/sentence embeddings for a document~\cite{sukhbaatar-15, kumar2016ask}.
To endow more structure and interpretability to memory, key-value memory networks were introduced by~\citet{miller-etal-2016-key}.
The key-value architecture has since been used for narrative understanding and other tasks where the memory is intended to learn to track entities %
while being guided by varying degrees of supervision~\cite{henaff2016tracking, liu-etal-2018-narrative, liu-etal-2018-recurrent, liu2019referential}.

We propose a new memory model, \modelname, for entity tracking and coreference resolution, inspired by the recent Referential Reader model~\cite{liu2019referential} but substantially simpler. %
Experiments on the GAP~\cite{webster2018gap} pronoun resolution task show that \modelname outperforms the Referential Reader with fewer parameters and simpler architecture.
Importantly, while Referential Reader performance degrades with larger memory, \modelname improves with increase in memory capacity (before saturation), which should enable tracking of a larger number of entities.
We conduct experiments to assess various memory architecture decisions, such as learning of memory initialization and separation of memory slots into key/value pairs.

To test interpretability of
memory models' entity tracking, we propose a new diagnostic evaluation based on entity counting---a task that the models are not explicitly trained for---using a small amount of annotated data.
Additionally, we conduct a small scale human evaluation to assess quality of people tracking based on model memory logs. \modelname substantially outperforms Referential Reader on both measures, indicating better and more interpretable tracking of people.\footnote{Code available at \url{https://github.com/shtoshni92/petra}}

\section{Model}

\begin{figure}[th]
  \centering
  \includegraphics[width=0.4\textwidth]{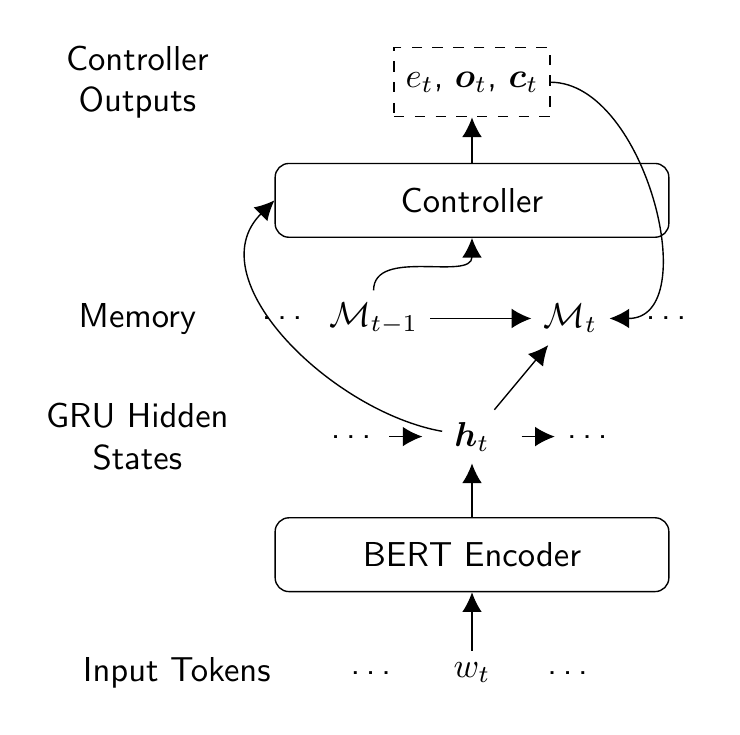}
  \captionof{figure}{Proposed model.}
  \label{fig:model_sch}
\end{figure}

Figure~\ref{fig:model_sch} depicts \modelname, which
consists of three components: an {\it input encoder} that given the tokens generates the token embeddings, a {\it memory module} that tracks information about the entities present in the text, and a {\it controller network} that acts as an interface between the encoder and the memory.

\subsection{Input Encoder}
\label{sec:input_encoder}
Given a document consisting of a sequence of tokens $\{w_1, \cdots, w_T\}$, we first pass the document through a fixed pretrained BERT model~\cite{devlin2019bert} to extract contextual token embeddings.
 Next, the BERT-based token embeddings are fed into a single-layer unidirectional Gated Recurrent Unit (GRU) ~\cite{cho2014learning} running left-to-right to get task-specific token embeddings $\{\vec{h}_1, \cdots, \vec{h}_T\}$.

\subsection{Memory}
The memory $\mathcal{M}_t$ consists of $N$ memory cells. The $i^{\textrm{th}}$ memory cell state at time step $t$ consists of a tuple $(\vec{m}^i_t, u^i_t)$
where the vector $\vec{m}^i_t$ represents the content of the memory cell, and the scalar $u^i_t \in [0, 1]$ represents
its recency of usage. %
A high value of $u^i_t$ is intended to mean that the cell is tracking an entity that has been recently mentioned. %

\paragraph{Initialization}
Memory cells are initialized to the null tuple, i.e.\ (\vec{0}, 0); thus, our memory is parameter-free. This is in contrast with previous %
entity tracking models such as EntNet~\cite{henaff2016tracking} and the Referential Reader~\cite{liu2019referential} where memory initialization is learned and the cells are represented with separate {\it key} and {\it value} vectors.
We will later discuss variants of our memory with some of these changes.

\subsection{Controller}
At each time step $t$ the controller network determines whether token $t$ is part of an entity span and, if so, whether the token is coreferent with any of the entities already being tracked by the memory. Depending on these two
variables, there are three possible actions: %
\begin{enumerate}[label=(\roman*)]
\itemsep0em
\item \actignore: The token is not part of any entity span, in which case we simply ignore it.
\item \actoverwrite: The token is part of an entity span \emph{but} is not already being tracked in the memory.
\item \actcoref: The token is part of an entity span and the entity is being tracked in the memory.
\end{enumerate}
Therefore, the two ways of updating the memory are \actoverwrite and \actcoref.
There is a strict ordering constraint to the two operations: \actoverwrite precedes \actcoref, because it is not possible to corefer with a memory cell that is not yet tracking anything. %
That is, the \actcoref operation cannot be applied to a previously unwritten memory cell, i.e.~one with $u^i_t = 0$. Figure~\ref{fig:ideal_run} illustrates an idealized version of this process.

Next we describe in detail the computation of the probabilities of the two operations for each memory cell at each time step $t$.

First, the {\bf entity mention probability} $e_t$, which reflects the probability that the current token $w_t$ is part of an entity mention, is computed by:
\begin{equation} \label{ent_eqn}
e_t = \sigma(\mlp_1(\vec{h}_t))
\end{equation}
where $\mlp_1$ is a multi-layer perceptron and $\sigma$ is the logistic function.

\paragraph{Overwrite and Coref}
If the current token $w_t$ is part of an entity mention, we need to determine whether it corresponds to an entity being currently tracked by the memory or not.
For this we compute the similarity between the token embedding $\vec{h}_t$ and the contents of the memory cells currently tracking entities.
For the $i^\textrm{th}$ memory cell with memory vector $\vec{m}_{t-1}^i$ the similarity with $\vec{h}_t$ is given by:%
\begin{equation} \label{sim_eqn}
\mysim_{t}^{i} = \mlp_2([\vec{h}_t; \vec{m}_{t-1}^{i};
                \vec{h}_t \odot \vec{m}_{t-1}^{i}; u_{t-1}^i])
\end{equation}
where $\mlp_2$ is a second MLP and $\odot$ is the Hadamard (elementwise) product.
The usage scalar $u_{t-1}^i$ in the above expression provides a notion of distance between the last mention of the entity in cell $i$ and the potential current mention.
The higher the value of $u_{t-1}^i$, the more likely
there was a recent mention of the entity being tracked by the cell.
Thus %
$u_{t-1}^i$ provides an alternative to distance-based features commonly used in pairwise scores for spans
\citep{lee-etal-2017-end}.

Given the entity mention probability $e_t$ and similarity score $\mysim_{t}^{i}$, we define the \textbf{coref score} $\cs_{t}^{i}$ as:
\begin{equation} \label{coref_score_eqn}
\cs_{t}^{i} = \mysim_{t}^{i} - \infty \cdot \mathds{1} [u^i_{t-1} = 0]
\end{equation}
where the second term ensures that the model does not predict coreference with a memory cell that has not been previously used, something not enforced by \citet{liu2019referential}.\footnote{A threshold higher than 0 can also be used to limit coreference to only more recent mentions.} Assuming the coref score for a new entity to be 0,\footnote{The new entity coref score is a free variable that can be assigned any value, since only the relative value %
matters.} we compute the \textbf{coref probability} $c_{t}^{i}$ and \textbf{new entity probability} $n_t$ as follows:
\begin{equation}
\label{coref_over_eqn}
          \begin{pmatrix}
           c_t^1 \\
           \vdots \\
           c_t^N \\
           n_t
          \end{pmatrix} = e_t \cdot \text{ softmax}
          \begin{pmatrix}
           \cs_{t}^{1} \\
           \vdots \\
           \cs_{t}^{N} \\
           0
         \end{pmatrix}
\end{equation}
Based on the memory usage scalars $u^i_t$ and the new entity probability $n_t$, the \textbf{overwrite probability} for each memory cell is determined as follows:
\begin{equation}\label{over_eqn_inf}
o_t^{i} = n_t \cdot \mathbbm{1}_{i = \arg\min_j u^j_{t-1}}
\end{equation}
Thus we pick the cell with the lowest usage scalar $u^j_{t-1}$ to \actoverwrite. In case of a tie, a cell is picked randomly among the ones with the lowest usage scalar.
The above operation is non-differentiable,
so during training we instead use %
\begin{equation}\label{over_eqn_train}
o_t^{i} = n_t \cdot \text{GS}\left(\frac{1 - u_{t-1}^{i}}{\tau}\right)_i
\end{equation}
where $\text{GS}(.)$ refers to Gumbel-Softmax~\cite{jang2017categorical}, which makes overwrites differentiable. %

For each memory cell, the memory vector is updated based on the three possibilities of ignoring the current token, being coreferent with the token, or considering the token to represent a new entity (causing an overwrite):
\begin{equation}
\begin{aligned}\label{memory_up}
    \vec{m}_t^{i}  = &\ \overbrace{(1 - (o_t^{i} + c_t^{i})) \vec{m}_{t-1}^{i}}^{\mbox{\actignore}}  \,+\!\!\!\!\!\! \overbrace{o_t^{i}\cdot \vec{h}_t}^{\mbox{\actoverwrite}} \\[3pt]
    & +\, \underbrace{c_t^{i} \cdot \mlp_3([\vec{h}_t; \vec{m}_{t-1}^i])}_{\mbox{\actcoref}}
\end{aligned}
\end{equation}
In this expression, the coreference term takes into account both the previous cell vector $\vec{m}_{t-1}^{i}$ and the current token representation $\vec{h}_t$, while the overwrite term is based only on $\vec{h}_t$.  In contrast to a similar
 memory update equation in the Referential Reader which employs a pair of GRUs and MLPs for each memory cell, our update parameter uses just $\mathrm{MLP}_3$ which is memory cell-agnostic. 

Finally, the memory usage scalar is
updated as
\begin{equation}
 u_t^{i} = \min(1, o_t^{i} + c_t^{i} + \gamma \cdot u_{t-1}^{i})
\end{equation}
where $\gamma \in (0, 1)$ is the decay rate for the usage scalar.
Thus the usage scalar $u_t^{i}$ keeps decaying with time unless the memory is updated via \actoverwrite or \actcoref in which case the value is increased to reflect the memory cell's recent use.

\paragraph{Memory Variants}
In vanilla \modelname, each memory cell is represented as a single vector and the memory is parameter-free,
so the total number of model parameters is independent of memory size.  This is a property that is shared with, for example, %
differentiable neural computers \citep{graves2016hybrid}.
On the other hand, recent models for entity tracking, such as the EntNet~\cite{henaff2016tracking} and the %
Referential Reader~\cite{liu2019referential}, learn memory initialization parameters and separate the memory cell into key-value pairs.
To compare these memory cell architectures, we investigate the following two variants of \modelname:
\begin{enumerate}
    \item \emph{\modelname + Learned Initialization}: memory cells are initialized at $t=0$ to learned parameter vectors.
    \item \emph{\modelname + Fixed Key}: a fixed dimensions of each memory cell are initialized with  learned parameters and kept fixed throughout the document read, as in EntNet~\cite{henaff2016tracking}.
\end{enumerate}
Apart from initialization, %
the initial cell vectors are also used to break ties for overwrites in Eqs.~\eqref{over_eqn_inf} and \eqref{over_eqn_train} when deciding among unused cells (with $u^i_t = 0$). The criterion for breaking the tie is the similarity score computed using Eq.~\eqref{sim_eqn}.

\subsection{Coreference Link Probability}
\label{sec:coref_link_prob}
The probability that the tokens $w_{t_1}$ and $w_{t_2}$ are coreferential according to, say, cell $i$ of the memory depends on three things:
\begin{enumerate*}[label=(\alph*)]
    \item $w_{t_1}$ is identified as part of an entity mention and is either overwritten to cell $i$ or is part of an earlier coreference chain for an entity tracked by cell $i$,
    \item Cell $i$ is not overwritten by any other entity mention from $t = {t_1} + 1$ to $t = {t_2}$, and
    \item $w_{t_2}$ is also predicted to be part of an entity mention and  is coreferential with cell $i$.
\end{enumerate*}
Combining these %
factors and marginalizing over the cell index results in the
following expression for the {\bf coreference link probability}: %
\begin{align}\label{prob_eqn}
P_{\mathrm{CL}}&(w_{t_1}, w_{t_2})  \nonumber\\
&= \sum_{i=1}^{N} (o_{t_1}^{i} + c_{t_1}^{i}) \cdot \! \prod_{j=t_1 + 1}^{t_2} (1 - o_{j} ^ {i}) \cdot c_{t_2}^{i}
\end{align}

\subsection{Losses}
The GAP~\cite{webster2018gap} training dataset is small and provides sparse supervision with labels for only two coreference links per instance.
In order to compensate for this lack of supervision, we use a heuristic loss $\mathcal{L}_{\mathit{ent}}$ over entity mention probabilities  %
in combination with the end task loss $\mathcal{L}_{\mathit{coref}}$ for coreference. The two losses are combined with a tunable hyperparameter $\lambda$ resulting in the following total loss: $\mathcal{L} = \mathcal{L}_{\mathit{coref}} + \lambda \mathcal{L}_{\mathit{ent}}$. 

\subsubsection{Coreference Loss}
\label{sec:coref_loss}
The coreference loss is the binary cross entropy between the ground truth labels for mention pairs and the coreference link probability $P_{\mathrm{CL}}$ in 
Eq.~\eqref{prob_eqn}. 
Eq.~\eqref{prob_eqn} expects 
a pair of tokens while the annotations are on pairs of spans, %
so we compute the loss for all ground truth token pairs: $\mathcal{L}_{\mathit{coref}} =$
\begin{align*}
    \sum_{(s_a, s_b, y_{ab}) \in {\mathrm{G}}} \left(\sum_{w_a \in s_a} \sum_{w_b \in s_b} H(y_{ab}, P_{\mathrm{CL}}(w_a, w_b))\right)
\end{align*}
where $\mathrm{G}$ is the set of annotated span pairs and $H(p, q)$ represents the cross entropy of the distribution $q$ relative to distribution $p$.

Apart from the ground truth labels, we use 
``implied labels" %
in the coreference loss calculation.
 For handling multi-token spans, we assume that all tokens following the head token are coreferential %
  with the head token (self-links). 
 We infer more supervision based on knowledge %
 of the %
 setup of the GAP task. Each GAP instance has two candidate names and a pronoun mention with supervision provided for the \{name, pronoun\} pairs. By design the two names are different, and therefore we use them as a negative coreference pair.
 
 Even after the addition of this implied %
 supervision, 
 our coreference loss calculation is restricted to the three mention spans in each training instance; therefore, the running time is $\mathcal{O}(T)$ for finite-sized mention spans. 
In contrast, \citet{liu2019referential} compute the above coreference loss for all token pairs (assuming a negative label for all %
pairs outside of the mentions), which results in a runtime of $\mathcal{O}(T^3)$ due to the   $\mathcal{O}(T^2)$ pairs and $\mathcal{O}(T)$ computation per pair, and thus will scale poorly to long documents.

\subsubsection{Entity Mention Loss}
\label{sec:ent_pred_loss}
We use the inductive bias that most tokens do not correspond to entities by imposing a loss on the average of the entity mention probabilities predicted across time steps, after masking out the labeled entity spans. 
For a %
training instance where spans $s_A$ and $s_B$ correspond to the %
person mentions and span $s_P$ is a pronoun, the entity mention loss is %
\vspace{-0.05in}
$$\mathcal{L}_{\mathit{ent}} = \frac{\sum_{t=1}^T e_t \cdot m_t}{\sum_{t=1}^T m_t}$$ 
where $m_t = 0$ if $w_t \in s_A \cup s_B \cup s_P$ and $m_t = 1$ otherwise. 

Each GAP instance has only 3 labeled entity mention spans, but the text typically has other entity mentions that are not labeled. 
Unlabeled entity mentions will be 
inhibited by this loss. However, on average there are far more tokens outside entity spans than inside the spans. 
In experiments without this loss, we observed that the model is susceptible to 
predicting a high entity probability for all tokens while still performing well on the end task of pronoun resolution. %
We are interested in tracking people beyond just the entities that are labeled in the GAP task, 
for which this loss is very helpful. %

\section{Experimental Setup}
\subsection{Data}
GAP is a gender-balanced pronoun resolution dataset
introduced by \citet{webster2018gap}.
Each instance consists of a small snippet of text from Wikipedia, two spans corresponding to candidate names along with a pronoun span, and two binary labels indicating the coreference relationship between the pronoun and the two candidate names. Relative to other popular coreference datasets~\cite{pradhan2012conll, chen-etal-2018-preco},
GAP is comparatively small and sparsely annotated. We choose GAP because its small size allows us to do extensive experiments.

\subsection{Model Details}
For the input BERT embeddings, we concatenate either the last four layers of \bertbase, %
or layers 19--22 of \bertlarge since those layers have been found to carry the most information related to coreference~\cite{liu2019linguistic}. The BERT embeddings are fed to a 300-dimensional GRU model, which matches the dimensionality of the memory vectors.

We vary the number of memory cells $N$
from 2 to 20. The decay rate for the memory usage scalar $\gamma$ is 0.98. The MLPs used for predicting the entity probability and similarity score consist of two 300-dimensional ReLU hidden layers.
For the \emph{Fixed Key} variant of \modelname we use 20 dimensions for the learned key vector and the remaining 280 dimensions as the value vector.

\begin{figure*}[ht]
\centering
\begin{subfigure}[b]{0.5\textwidth}
        \centering
        \includegraphics[width=\textwidth]{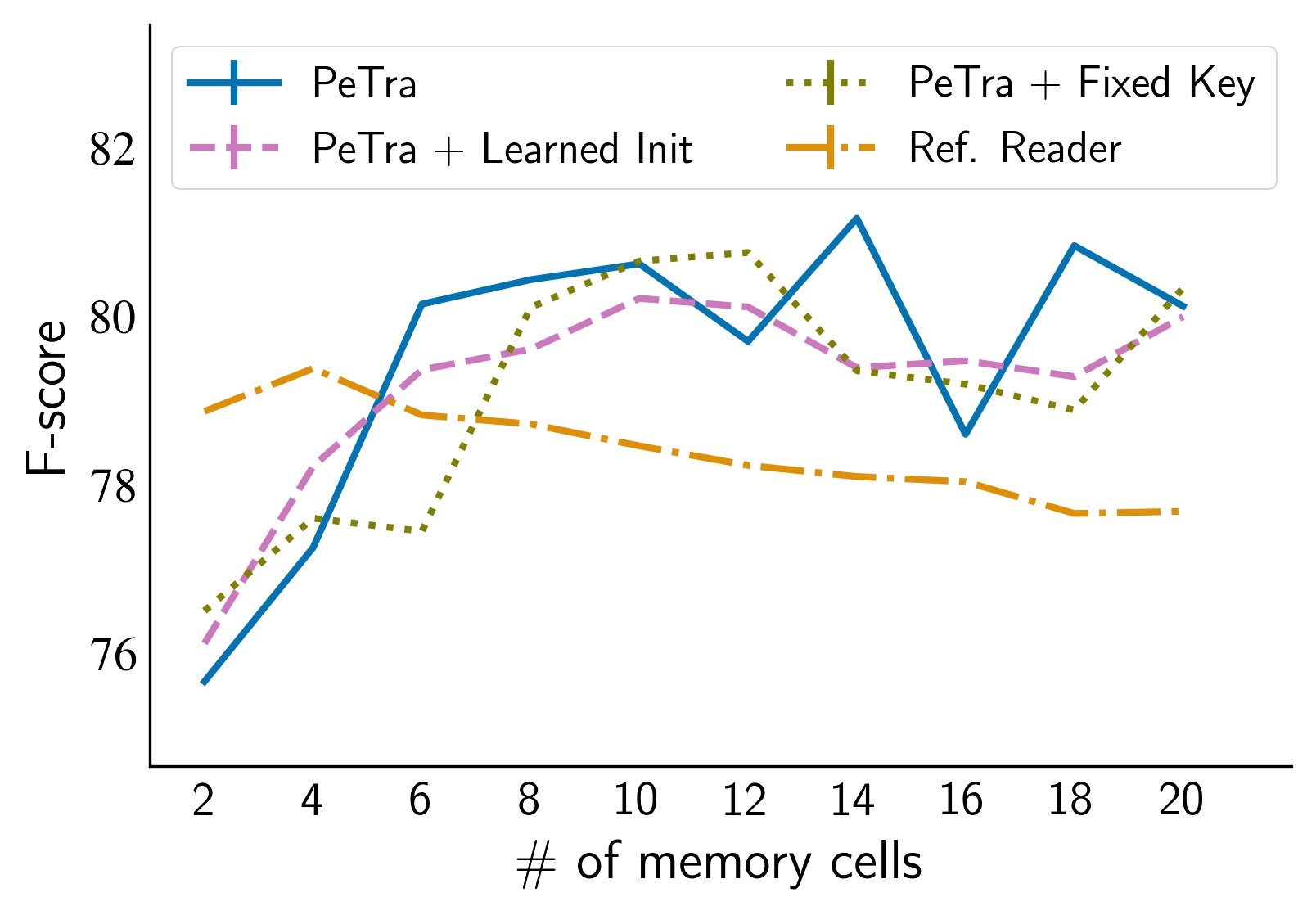}
        \caption{\bertbase}
        \label{fig:gap_val_small}
    \end{subfigure}%
    ~
    \begin{subfigure}[b]{0.5\textwidth}
        \centering
        \includegraphics[width=\textwidth]{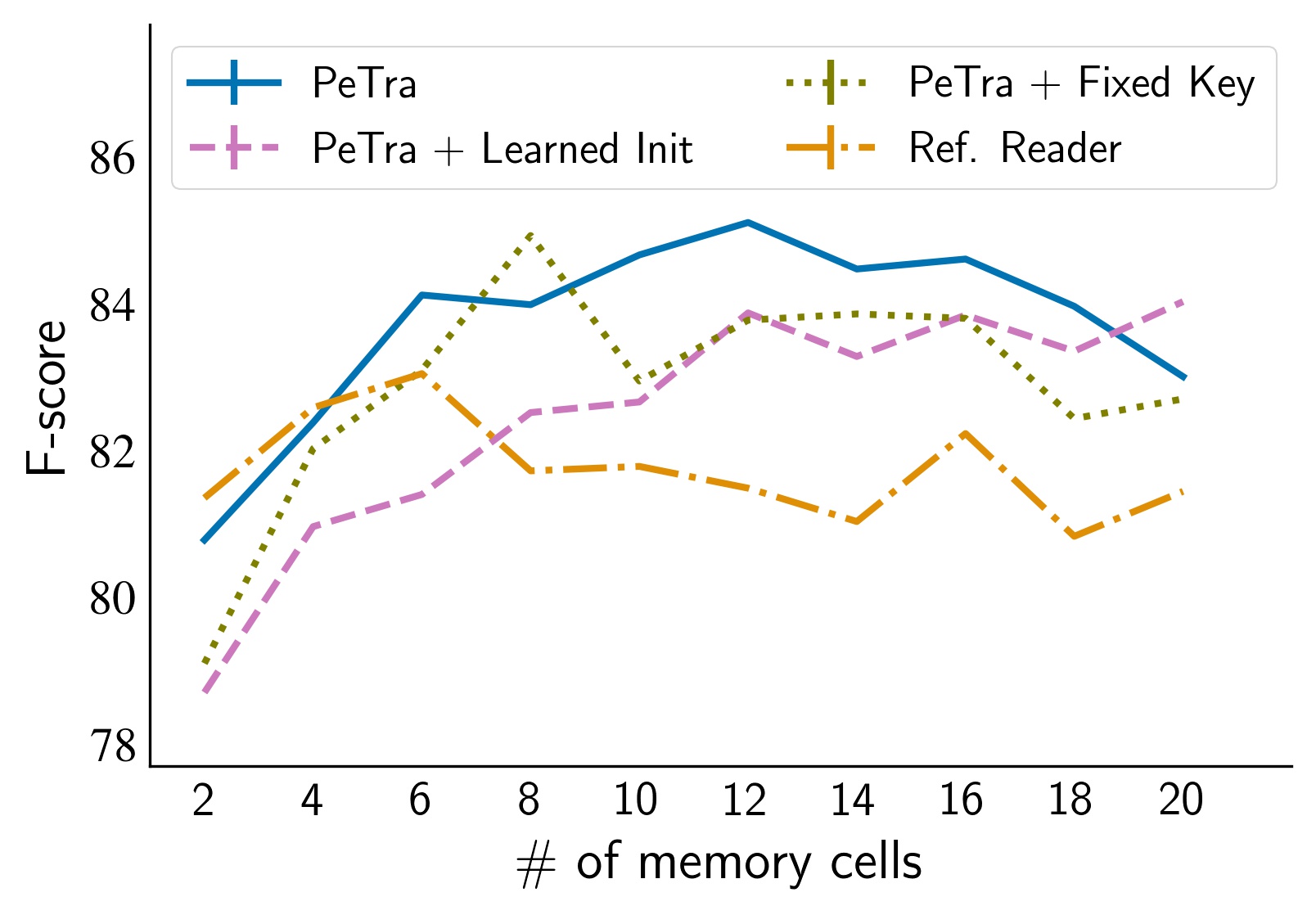}
        \caption{\bertlarge}
        \label{fig:gap_val_large}
    \end{subfigure}
    \caption{Mean F1 score on the GAP validation set as a function of the number of memory cells.}
    \label{fig:gap_val}
\end{figure*}

\subsection{Training}
All models are trained for a maximum of 100 epochs with the Adam optimizer~\cite{Kingma2015AdamAM}.
The learning rate is initialized to $10^{-3}$
and is reduced by half, until a minimum of $10^{-4}$,
whenever there is no improvement on the validation performance for the last 5 epochs.
Training stops when there is no improvement in validation performance for the last 15 epochs.
The temperature $\tau$ of the Gumbel-Softmax distribution used in the \actoverwrite operation is initialized to $1$ and halved every 10 epochs.
The coreference loss terms in Section~\ref{sec:coref_loss} are weighted differently for different coreference links: \begin{enumerate*}[label=(\alph*)]
    \item self-link losses for multi-token spans are given a weight of 1,
    \item positive coreference link losses are weighted by 5, and
    \item negative coreference link losses are multiplied by 50.
\end{enumerate*}
To prevent overfitting: \begin{enumerate*}[label=(\alph*)]
    \item we use early stopping based on validation performance, and
    \item apply dropout at a rate of 0.5 on the output of the GRU model.
\end{enumerate*}
Finally, we choose $\lambda=0.1$ to weight the entity prediction loss described in Section~\ref{sec:ent_pred_loss}.

\subsection{People Tracking Evaluation}
\label{sec:exp_people_tracking_eval}
One of the goals of this work is to develop memory models that not only do well on the coreference resolution task, but also are interpretable in the sense that the memory cells actually track entities. Hence in addition to reporting the standard metrics on GAP, we consider two other ways to evaluate memory models. %

  As our first task, we propose an auxiliary entity-counting task. %
We take 100 examples from the GAP validation set and annotate them
with the number of unique people mentioned in them.\footnote{In the GAP dataset, the only relevant entities are people.}
We test the models by predicting the number of people from their memory logs as explained in Section~\ref{sec:inference}.
The motivation behind this exercise is that if a memory model is truly tracking entities, then its memory usage logs should allow us to recover  this information.

To assess the people tracking performance more holistically, we conduct a human evaluation in which we ask annotators to assess the memory models on people tracking performance, defined as:%
(a) detecting references to people including pronouns, and (b) maintaining a 1-to-1 correspondence between people and memory cells.
For this study, we pick the best run (among 5 runs) of \modelname and the Referential Reader for the 8-cell configuration using \bertbase (\modelname: 81 F1; Referential Reader: 79 F1).
Next we randomly pick 50 documents (without replacement) from the GAP dev set and split those into groups of 10 to get 5 evaluation sets.
We shuffle the original 50 documents and follow the same steps to get another 5 evaluation sets.
In the end, we have a total of 10 evaluation sets with 10 documents each, where each unique document belongs to exactly 2 evaluation sets.

We recruit 10 annotators for the 10 evaluation sets.
The annotators are shown memory log visualizations as in Figure~\ref{fig:visualize}, and instructed to compare the models on their people tracking performance (detailed instructions in Appendix~\ref{sec:app_hum_eval}). %
For each document the annotators are presented memory logs of the two models (ordered randomly) and asked whether they prefer the first model, prefer the second model, or have no preference
(neutral). %

\subsection{Inference}
\label{sec:inference}
\paragraph{GAP} %
Given a pronoun span $s_P$ and two candidate name spans $s_A$ \&  $s_B$, we
have to predict binary labels for potential coreference links between ($s_A$, $s_P$) and ($s_B$, $s_P$). %
Thus, for a pair of entity spans, say $s_A$ and $s_P$, we predict the coreference link probability as:
\begin{align*}
    P_{\mathrm{CL}}(s_A, s_P) = \max_{w_A \in s_A, w_P \in s_P} P_{\mathrm{CL}}(w_A, w_P)
\end{align*}
where $P_{\mathrm{CL}}(w_A, w_P)$ is calculated using the procedure described in Section~\ref{sec:coref_link_prob}\footnote{The computation of this probability includes the mention detection steps required by\citet{webster2018gap}.}. The final binary prediction is made by comparing the probability against a threshold.

\begin{figure*}[t]
\begin{subfigure}[b]{0.5\textwidth}
        \centering
        \includegraphics[width=\textwidth]{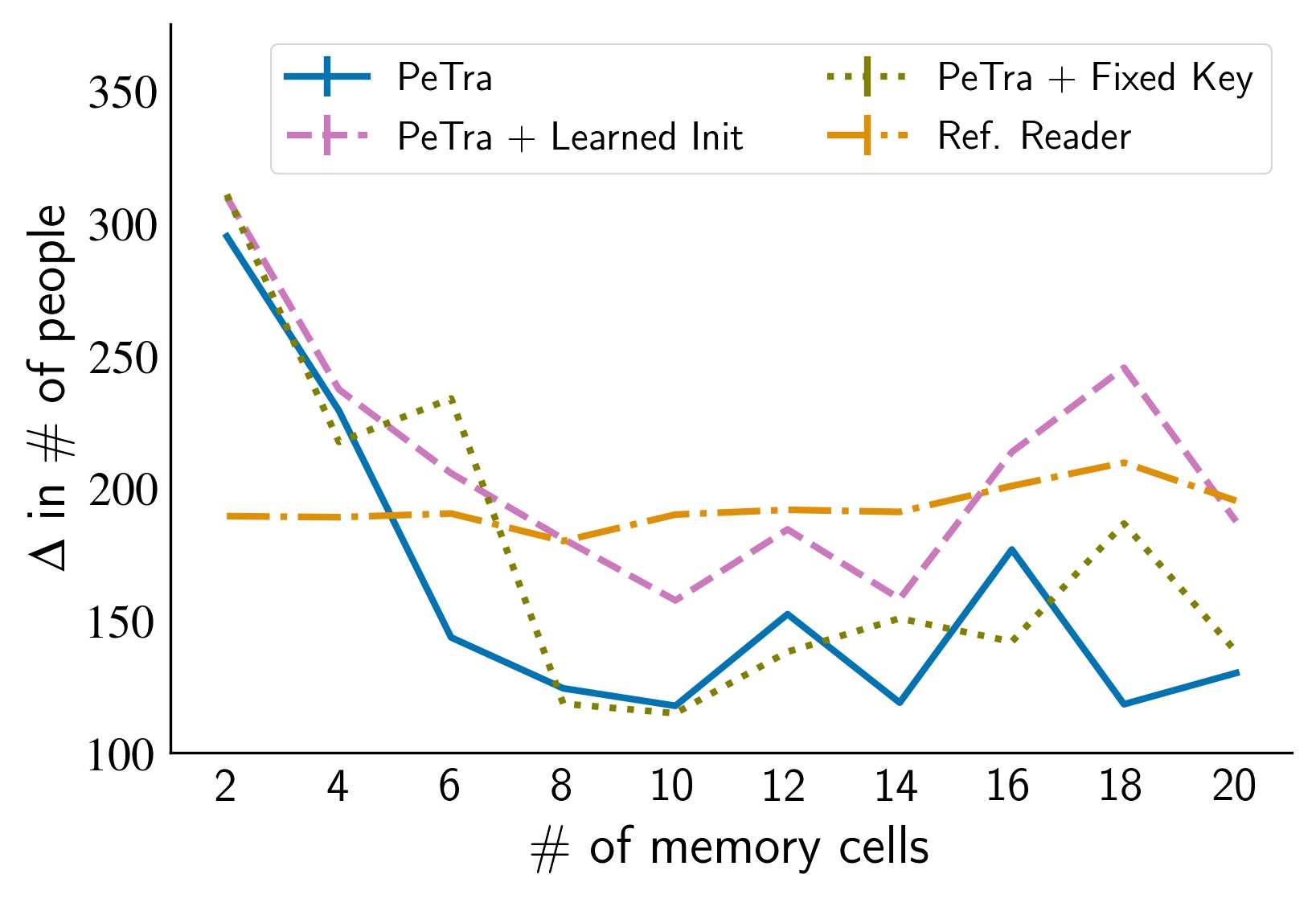}
        \caption{\bertbase}
        \label{fig:count_uniq_people_small}
    \end{subfigure}%
    ~
    \begin{subfigure}[b]{0.5\textwidth}
        \centering
        \includegraphics[width=\textwidth]{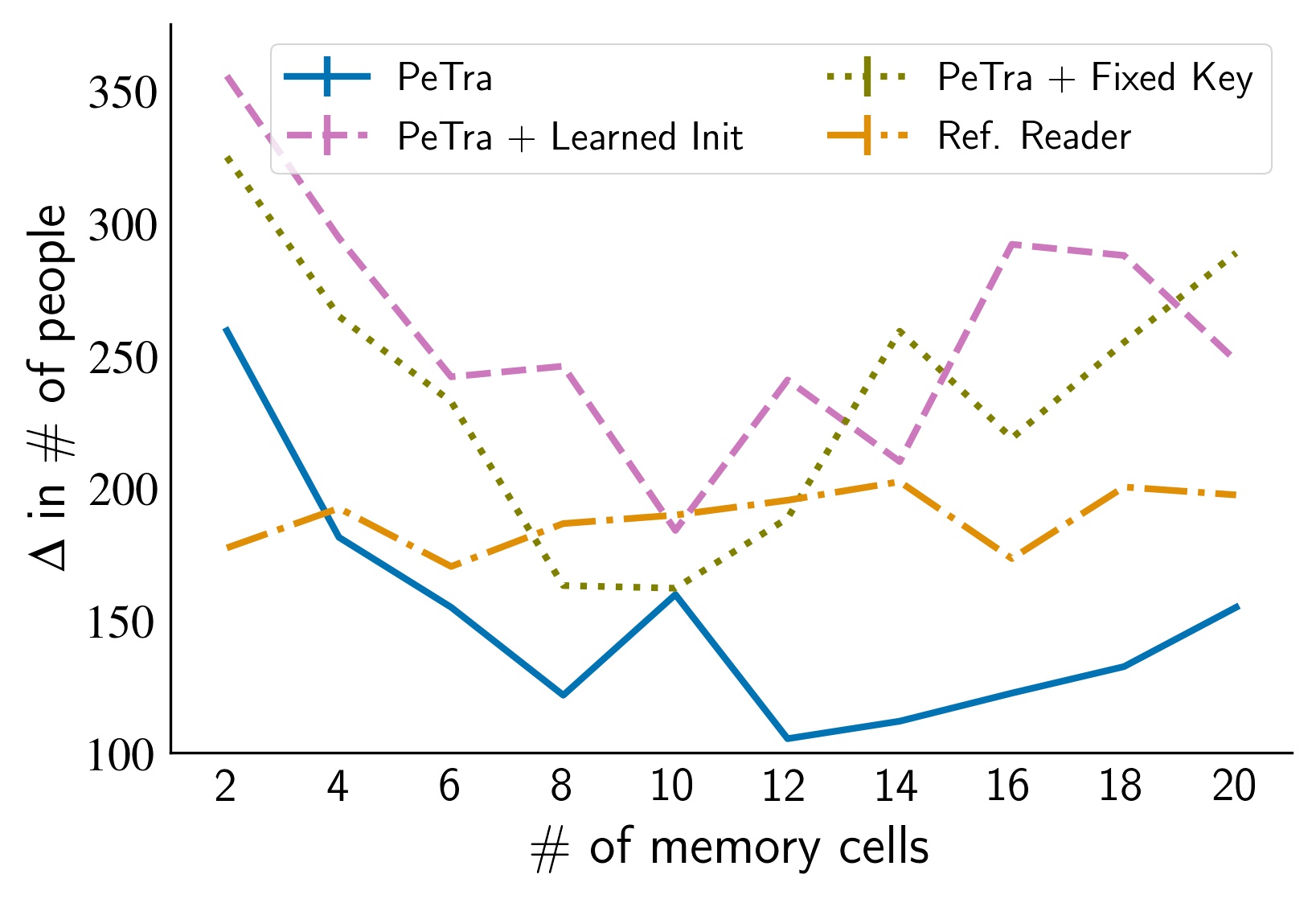}
        \caption{\bertlarge}
        \label{fig:count_uniq_people_large}
    \end{subfigure}
    \caption{Error in counting unique people as a function of number of memory cells; lower is better. }
    \label{fig:count_uniq_people}
    \vspace{-0.1in}
\end{figure*}

\paragraph{Counting unique people}
For the test of unique
people counting, we discretize the overwrite operation, which corresponds to new entities, against a threshold $\alpha$ and sum over all
tokens and all
memory cells to predict the count as follows:\vspace{-0.02in}
$$\text{\# unique people} = \sum_{t=1}^T\sum_{i=1}^N \mathbbm{1}[o^i_t \geq \alpha] $$

\subsection{Evaluation Metrics}
For GAP we evaluate models using F-score.\footnote{GAP also includes evaluation related to gender bias, but this is not a focus of this paper so we do not report it.}
First, we pick a threshold from the set \{0.01, 0.02, $\cdots$, 1.00\} which maximizes the validation F-score.
This threshold is then used to evaluate performance on the GAP test set.

For the interpretability task of counting unique people, we
choose a threshold that minimizes the absolute difference between ground truth count and predicted count summed over the 100 annotated examples.
We select the best threshold from the set \{0.01, 0.02, $\cdots$, 1.00\}.
The metric is then the number of errors corresponding to the best threshold.\footnote{Note that the error we report is therefore a best-case result.  We are not proposing a way of counting unique people in new test data, but rather using this task for analysis.}

\subsection{Baselines}
The Referential Reader~\cite{liu2019referential} is the most relevant baseline in the literature, and the most similar to \modelname.
The numbers reported by~\citet{liu2019referential} are obtained by a version of the model using \bertbase, with only two memory cells.
To compare against \modelname for other configurations, we retrain the Referential Reader using the code made available by the authors.\footnote{\url{https://github.com/liufly/refreader}}

We also report the results of \citet{joshi-etal-2019-bert} and \citet{wu2019coreference}, %
although these numbers are not comparable since both of them train on the much larger OntoNotes corpus and just test on GAP.

\section{Results}

\begin{figure*}[!ht]
    \centering

    \begin{subfigure}[t]{\textwidth}
    \begin{mdframed}
        \footnotesize{
        \hlent{Amelia Shepherd}{1}, M.D. is a fictional character on the ABC American television medical drama Private Practice, and the spinoff series' progenitor show, Grey's Anatomy, portrayed by \hlent{\textcolor{red}{\it Caterina Scorsone}}{2}. In \hlent{\textcolor{aqua}{\bf her}}{1} debut appearance in season three, \hlent{\textcolor{aqua}{\bf Amelia}}{1} visited her former sister-in-law, \hlent{Addison Montgomery}{3}, and became a partner at the Oceanside Wellness Group.
        }
    \end{mdframed}
    \end{subfigure}
    \begin{subfigure}[t]{\textwidth}
    \includegraphics[width=\textwidth]{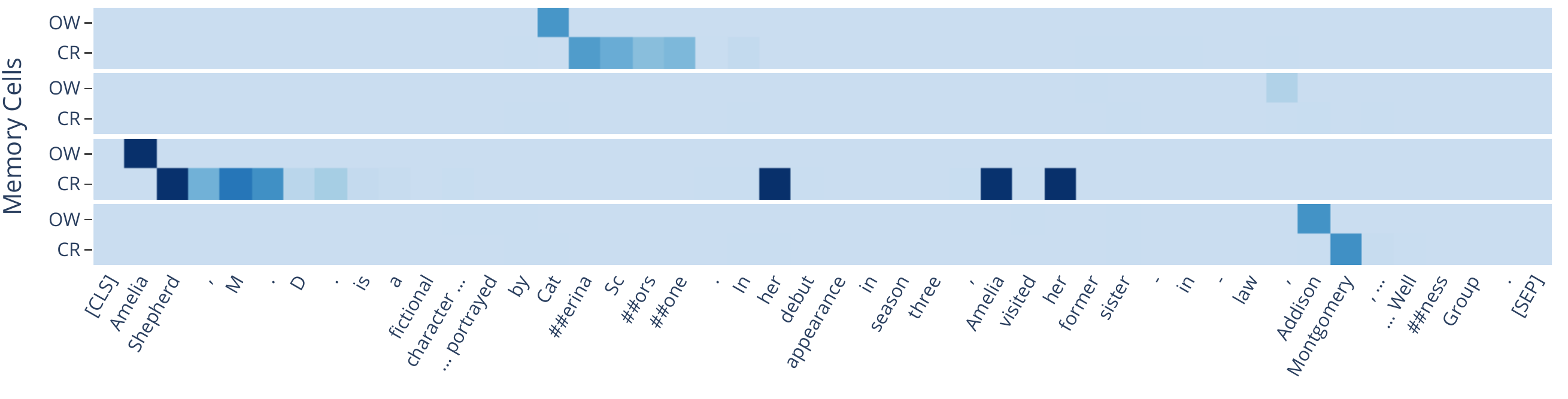}
    \caption{A successful run of \modelname with 4 memory cells. The model accurately links all the mentions of ``Amelia" to the same memory cell while also detecting other people in the discourse.}
    \label{fig:success}
    \end{subfigure}
    \begin{subfigure}[t]{\textwidth}
    \begin{mdframed}
        \footnotesize{
        \hlent{Bethenny}{1} calls a meeting to get everyone on the same page, but \hlent{Jason}{2} is hostile with the group, making things worse and forcing \hlent{Bethenny}{1} to play referee. Emotions are running high with \hlent{\textcolor{red}{\it Bethenny}}{1}'s assistant, \hlent{\textcolor{aqua}{\bf Julie}}{3}, who breaks down at a lunch meeting when asked if \hlent{\textcolor{aqua}{\bf she}}{3} is committed to the company for the long haul.
        }
    \end{mdframed}
    \end{subfigure}
    \begin{subfigure}[t]{\textwidth}
    \includegraphics[width=\textwidth]{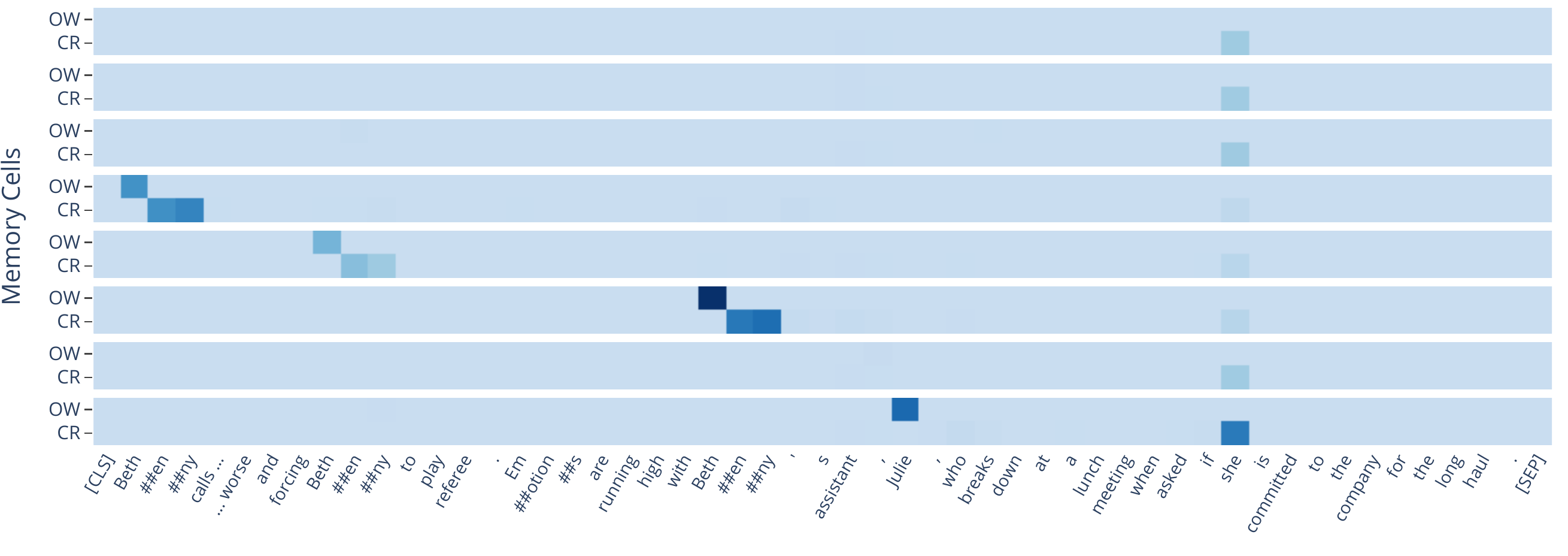}
    \caption{Memory log of \modelname with 8 memory cells. The model correctly links ``she" and ``Julie" but fails at linking the three ``Bethenny" mentions, and also fails at detecting ``Jason".}
    \label{fig:failure}
    \end{subfigure}
    \caption{Visualization of memory logs for different configurations of \modelname.
    The documents have their GAP annotations highlighted in red (italics) and blue (bold), with blue (bold) corresponding to the right answer. For illustration purposes only, we highlight all the spans corresponding to mentions of people and mark cluster indices as subscript.
    In the plot, X-axis corresponds to document tokens, and Y-axis corresponds to memory cells. Each memory cell has the OW=\actoverwrite and CR=\actcoref labels. Darker color implies higher value.
    We skip text, indicated via ellipsis, when the model doesn't detect people for extended lengths of text.}
    \label{fig:visualize}
\end{figure*}

\subsection{GAP results}
\begin{table}[h]
\centering{
\small{
\begin{tabular}{lll}
\toprule
 & \bertbase  & \bertlarge \\\midrule
\modelname                   & {\bf 81.5 $\pm$ 0.6} &  {\bf 85.3 $\pm$ 0.6} \\
\hspace{0.1in} + Learned Init. & 80.9 $\pm$ 0.7 &  84.4 $\pm$ 1.2\\
\hspace{0.1in} + Fixed Key & 81.1 $\pm$ 0.7 & 85.1 $\pm$ 0.8\\
Ref. Reader & 78.9 $\pm$ 1.3 & 83.7 $\pm$ 0.8\\
Ref. Reader ~\shortcite{liu2019referential} & 78.8 & - \\\midrule
\citet{joshi-etal-2019-bert} & 82.8 & 85.0 \\
\citet{wu2019coreference} & - & 87.5 (SpanBERT)\\
\bottomrule
\end{tabular}
}
}
\caption{Results (\%F1) on the GAP test set.}
\label{tab:gap_res}
\end{table}

We train all the memory models, including the Referential Reader, with memory size varying from \{2, 4, $\cdots$, 20\} memory cells for both \bertbase and \bertlarge, with each configuration being trained 5 times.
Figure~\ref{fig:gap_val} shows the performance of the
models on the GAP validation set as a function of memory size.
The Referential Reader outperforms
\modelname (and its memory variants) when using a small number of memory cells, but its performance starts degrading after 4 and 6 memory cells for \bertbase and \bertlarge respectively. \modelname and its memory variants, in contrast, keep improving with increased memory size (before saturation at a higher number of cells) and outperform the best Referential Reader performance for all memory sizes $\ge$ 6 cells. With larger numbers of memory cells, we see a higher variance, but the curves for \modelname and its memory variants are still consistently higher than those of the Referential Reader.

Among different memory variants of \modelname  %
, when using \bertbase the performances are comparable with no clear advantage for any particular choice. %
For \bertlarge, however, vanilla \modelname has a clear edge for almost all memory sizes, suggesting the limited utility of initialization.
The results show that \modelname works well without learning vectors for initializing the key or memory cell contents. Rather, we can remove the key/value distinction and simply initialize all memory cells with the zero vector.

To evaluate on the GAP test set, we pick the memory size corresponding to the best validation performance for all memory models. Table~\ref{tab:gap_res} shows that the trends from validation hold true for test as well, with \modelname outperforming the Referential Reader and the other memory variants of \modelname.

\subsection{Counting unique people}
Figure~\ref{fig:count_uniq_people} shows the results for the proposed interpretability task of counting unique people.
For both \bertbase and \bertlarge, \modelname achieves the lowest error count.
Interestingly, from Figure~\ref{fig:count_uniq_people_large} we can see that for $\ge 14$ memory cells, the other memory variants of \modelname perform worse than the Referential Reader while being better at the GAP validation task (see Figure~\ref{fig:gap_val_large}). This shows that a better performing model is not necessarily better at tracking people.

\begin{table}[h]
\setlength{\tabcolsep}{5pt}
\small
\centering{
    \begin{tabular}{lcc}
    \toprule
         & \bertbase & \bertlarge \\
        \toprule
        \modelname & \textbf{0.76} & \textbf{0.69} \\
        \hspace{0.1in} + Learned Init & 0.72 & 0.60 \\
        \hspace{0.1in} + Fixed Key & 0.72 & 0.65 \\
        Ref. Reader & 0.49 &  0.54 \\
        \bottomrule
    \end{tabular}
}
\caption{Spearman's correlation between GAP validation F1 and negative error count for unique people.} %
\label{tab:correlation}
\vspace{-0.1in}
\end{table}
To test the relationship between the GAP task and the proposed interpretability task, we compute the correlation between the GAP F-score and the negative count of unique people for each model separately.\footnote{Each correlation is computed over 50 runs (5 runs each for 10 memory sizes).}
Table \ref{tab:correlation} shows the Spearman's correlation between these measures. For all models we see a positive correlation, indicating that a dip in coreference performance corresponds to an increase in error on counting unique people. The correlations for \modelname are especially high, again suggesting it's greater interpretability.

\subsection{Human Evaluation for People Tracking}
\label{sec:hum_eval}

\begin{table}[h]
\centering{
\small{
\begin{tabular}{lc}
\toprule
Model & Preference (in \%) \\
\midrule
\modelname        & \textbf{74} \\
Ref. Reader &  08\\
Neutral & 18 \\\bottomrule
\end{tabular}
}
}
\caption{Human Evaluation results for people tracking.}
\label{tab:hum_eval}
\vspace{-0.1in}
\end{table}

Table~\ref{tab:hum_eval} summarizes the results of the human evaluation for people tracking. %
The annotators prefer \modelname in 74\% cases while the Referential Reader for only 8\% instances (see Appendix~\ref{sec:app_visualizations} for visualizations comparing the two). %
Thus, \modelname easily outperforms the Referential Reader on this task even though they are quite close on the GAP evaluation.
The annotators agree on 68\% of the documents, disagree between \modelname and Neutral for 24\% of the documents, and disagree between \modelname and the Referential Reader for  the remaining 8\%    documents. For more details, see Appendix~\ref{sec:app_hum_eval_res}.

\subsection{Model Runs}
We visualize two runs of \modelname with different configurations in Figure~\ref{fig:visualize}. For both instances the model gets the right pronoun resolution, but clearly in Figure~\ref{fig:failure} the model fails at correctly tracking repeated mentions of ``Bethenny". We believe these errors happen because (a) GAP supervision is limited to pronoun-proper name pairs, so the model is never explicitly supervised to link proper names, and (b) there is a lack of span-level features, which hurts the model when a name is split across multiple tokens. %

\section{Related Work}

There are several strands of related work, including prior work in developing neural models with external memory as well as variants that focus on modeling entities and entity relations, and neural models for coreference resolution.

\paragraph{Memory-augmented models.}

Neural network architectures with external memory include memory networks~\citep{weston-15,sukhbaatar-15}, neural Turing machines~\citep{graves2014neural}, and differentiable neural computers~\citep{graves2016hybrid}. %
This paper focuses on models with inductive biases that produce particular structures in the memory, specifically those related to entities.

\paragraph{Models for tracking and relating entities.}
A number of existing models have targeted entity tracking and coreference links for a variety of tasks.
EntNet~\citep{henaff2016tracking} aims to track entities via a memory model.
EntityNLM~\citep{D17-1195} represents entities dynamically within a neural language model. %
\citet{hoang-etal-2018-entity} augment a reading comprehension model to track entities, incorporating a set of %
 auxiliary losses to encourage capturing of reference relations in the text. %
\citet{dhingra-etal-2018-neural} introduce a modified GRU layer designed to aggregate information across coreferent mentions.%

\paragraph{Memory models for NLP tasks.}
Memory models have been applied to several other NLP tasks in addition to coreference resolution, including targeted aspect-based sentiment analysis~\citep{liu-etal-2018-recurrent}, machine translation~\citep{maruf-haffari-2018-document}, narrative modeling~\citep{liu-etal-2018-narrative}, and dialog state tracking~\citep{perez-liu-2017-dialog}. Our study of architectural choices for memory may also be relevant to models for these tasks.

\paragraph{Neural models for coreference resolution.}
Several neural models have been developed for coreference resolution, most of them focused on modeling pairwise interactions among mentions or spans in a document~\citep{wiseman-etal-2015-learning,clark-manning-2016-deep,lee-etal-2017-end,lee-etal-2018-higher}. These models use heuristics to avoid computing scores for all possible span pairs in a document, an operation which is quadratic in the document length $T$ assuming a maximum span length.
Memory models for coreference resolution, including our model, differ by seeking to store information about entities in memory cells and then modeling the relationship between a token and a memory cell. This reduces computation from $\mathcal{O}(T^2)$ to $\mathcal{O}(TN)$, where $N$ is the number of memory cells, allowing memory models to be applied to longer texts by using the global entity information. Past work ~\citep{wiseman-etal-2016-learning} have used global features, but in conjunction with other features to score span pairs.

\paragraph{Referential Reader.} Most closely related to the present work is the Referential Reader~\citep{liu2019referential}, which uses a memory model to perform coreference resolution incrementally. %
We significantly simplify this model to accomplish the same goal with far fewer parameters.

\section{Conclusion and Future Work}
We propose a new memory model for entity tracking, which is trained using sparse coreference resolution supervision.
The proposed model outperforms a previous approach with far fewer parameters and a simpler architecture.
We propose a new diagnostic evaluation and conduct a human evaluation to test the interpretability of the model, and find that
our model again does better on this evaluation.
In future work, we plan to extend this work to longer documents such as the recently released dataset of~\citet{bamman2019annotated}.

\section*{Acknowledgments}
{\footnotesize
This material is based upon work supported by the National Science Foundation under Award Nos.~1941178 and 1941160.
We thank the ACL reviewers, Sam Wiseman, and Mrinmaya Sachan for their valuable feedback.
We thank Fei Liu and Jacob Eisenstein for answering questions regarding the Referential Reader.
Finally, we want to thank all the annotators at TTIC who participated in the human evaluation study.%
}

\bibliography{acl2020}
\bibliographystyle{acl_natbib}
\appendix
\section{Appendix}

\subsection{Best Runs vs.~Worst Runs}
As Table~\ref{tab:gap_res} shows, there is significant variance in the performance of these memory models.
To analyze how the best runs diverge from the worst runs, we analyze how the controller network is using the different memory cells in terms of overwrites.
For this analysis, we choose the best and worst among the 5 runs for each configuration, as determined by GAP validation performance. For the selected runs, we calculate the KL-divergence of the average overwrite probability distribution from the uniform distribution and average it for each model type. Table~\ref{tab:best_vs_worst} shows that for the memory variants {\it Learned Init} and {\it Fixed Key}, the worst runs overwrite more to some memory cells than others (high average KL-divergence). Note that both \modelname and Referential Reader are by design intended to have no preference for any particular memory cell (which the numbers support), hence the low KL-divergence.
\begin{table}[h]
\small
\centering{
\begin{tabular}{lll}
\toprule
 & \multicolumn{2}{c}{Avg KL-div} \\
 & Best run & Worst run \\\midrule
\modelname & 0.00 & 0.01 \\
\hspace{0.1in} + Learned Init. & 0.3 & {\bf 0.83} \\
\hspace{0.1in} + Fixed Key & 0.2 & {\bf 0.8} \\
Ref. Reader & 0.05 & 0.04 \\
\bottomrule
\end{tabular}
}
\caption{A comparison of best runs vs.~worst runs. %
}
\label{tab:best_vs_worst}
\vspace{-0.1in}
\end{table}

\subsection{Human Evaluation Results}
\label{sec:app_hum_eval_res}
\begin{figure}[h]
    \centering
    \includegraphics[width=0.45\textwidth]{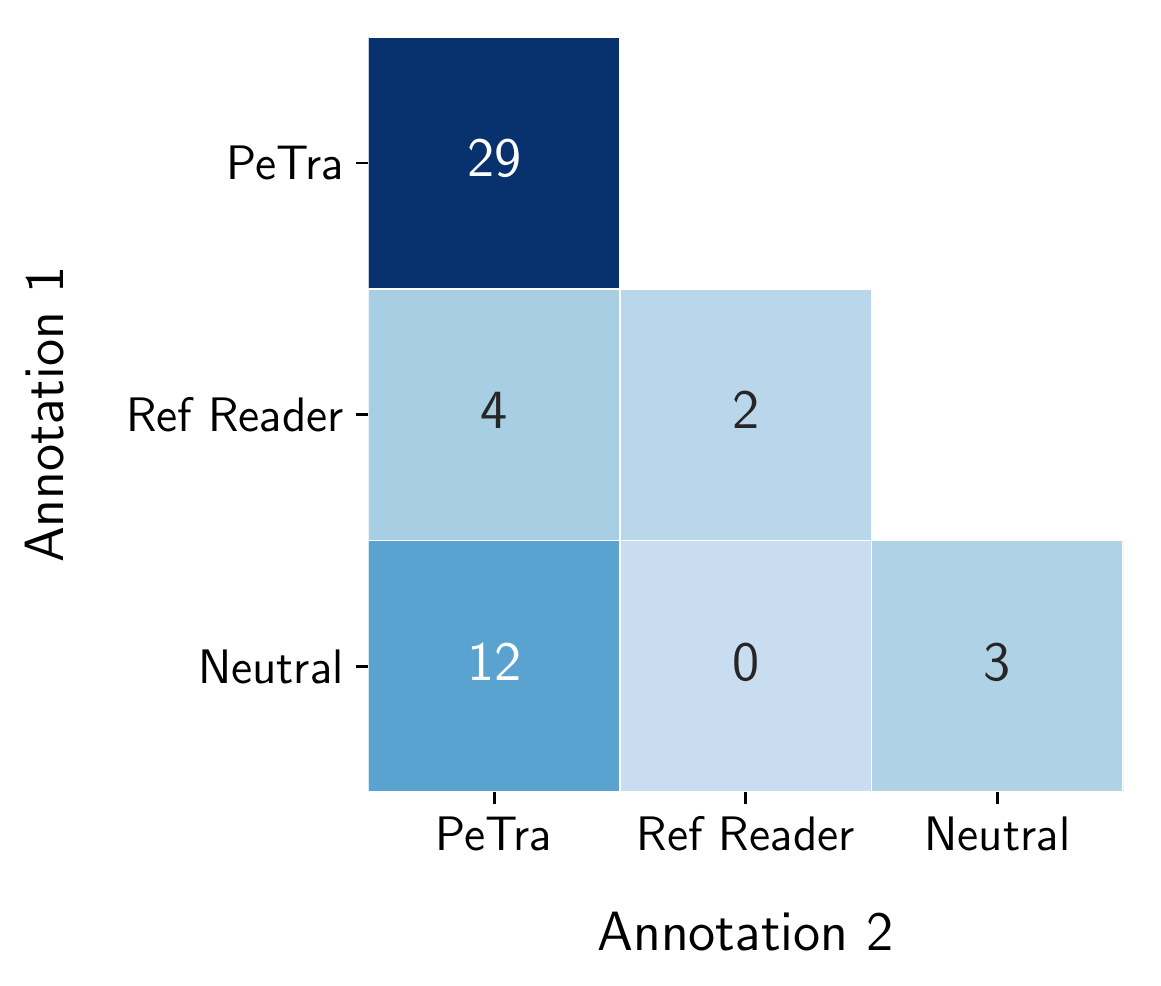}
    \caption{Agreement matrix for human evaluation study.}
    \label{fig:agreement}
\end{figure}

The agreement matrix for the human evaluation study described in Section~\ref{sec:hum_eval} is shown in Figure~\ref{fig:agreement}.
This agreement matrix is a result of the two annotations per document that we get as per the setup described in Section~\ref{sec:exp_people_tracking_eval}.
Note that the annotations are coming from two sets of annotators rather than two individual annotators. This is also the reason why we don't report standard inter-annotator agreement coefficients.

\subsection{Instructions for Human Evaluation}
The detailed instructions for the human evaluation study described in Section~\ref{sec:hum_eval} are shown in Figure~\ref{fig:instructions}.
We simplified certain memory model specific terms such as ``overwrite" to ``new person" since the study was really about people tracking.
\label{sec:app_hum_eval}
\begin{figure*}[ht]
\begin{mdframed}%
    \begin{itemize}
    \item In this user study we will be comparing memory models at tracking people.
    \item What are memory models? Memory models are neural networks coupled with an external memory which can be used for reading/writing.
    \item \textcolor{red}{(IMPORTANT)} What does it mean to track people for memory models?
    \begin{itemize}
        \item Detect all references to people which includes pronouns.
        \item A 1-to-1 correspondence between people and memory cells i.e. all references corresponding to a person should be associated with the same memory cell AND each memory cell should be associated with at most 1 person.
    \end{itemize}

    \item The memory models use the following scores (which are visualized) to indicate the tracking decisions:
    \begin{itemize}
    \item New Person Probability (Cell $i$): Probability that the token refers to a new person (not introduced in the text till now) and we start tracking it in cell $i$.
    \item Coreference Probability (Cell $i$): Probability that the token refers to a person already being tracked in cell $i$.
    \end{itemize}
    \item The objective of this study is to compare the models on the interpretability of their memory logs i.e.\ are the models actually tracking entities or not. You can choose how you weigh the different requirements for tracking people (from 3).
    \item For this study, you will compare two memory models with 8 memory cells (represented via 8 rows). The models are ordered randomly for each instance.
    \item For each document, you can choose model A or model B, or stay neutral in case both the models perform similarly.
    \end{itemize}
\end{mdframed}
\caption{Instructions for the human evaluation study.}
\label{fig:instructions}
\end{figure*}

\subsection{Comparative visualization of memory logs of \modelname and the Referential Reader}
\label{sec:app_visualizations}
\begin{figure*}[h]
    \begin{subfigure}[t]{\textwidth}
    \begin{mdframed}
        \hlent{\textcolor{aqua}{\bf Neef}}{1} took an individual silver medal at the 1994 European Cup behind Russia's \hlent{\textcolor{red}{\it Svetlana Goncharenko}}{2} and returned the following year to win gold. \hlent{\textcolor{aqua}{\bf She}}{1} was a finalist individually at the 1994 European Championships and came sixth for Scotland at the 1994 Commonwealth Games.
    \end{mdframed}
    \caption{GAP validation instance 293. The ground truth GAP annotation is indicated via colors.}
    \end{subfigure}
    ~\vspace{0.2in}
    \centering
    \begin{subfigure}[t]{\textwidth}
    \includegraphics[width=\textwidth]{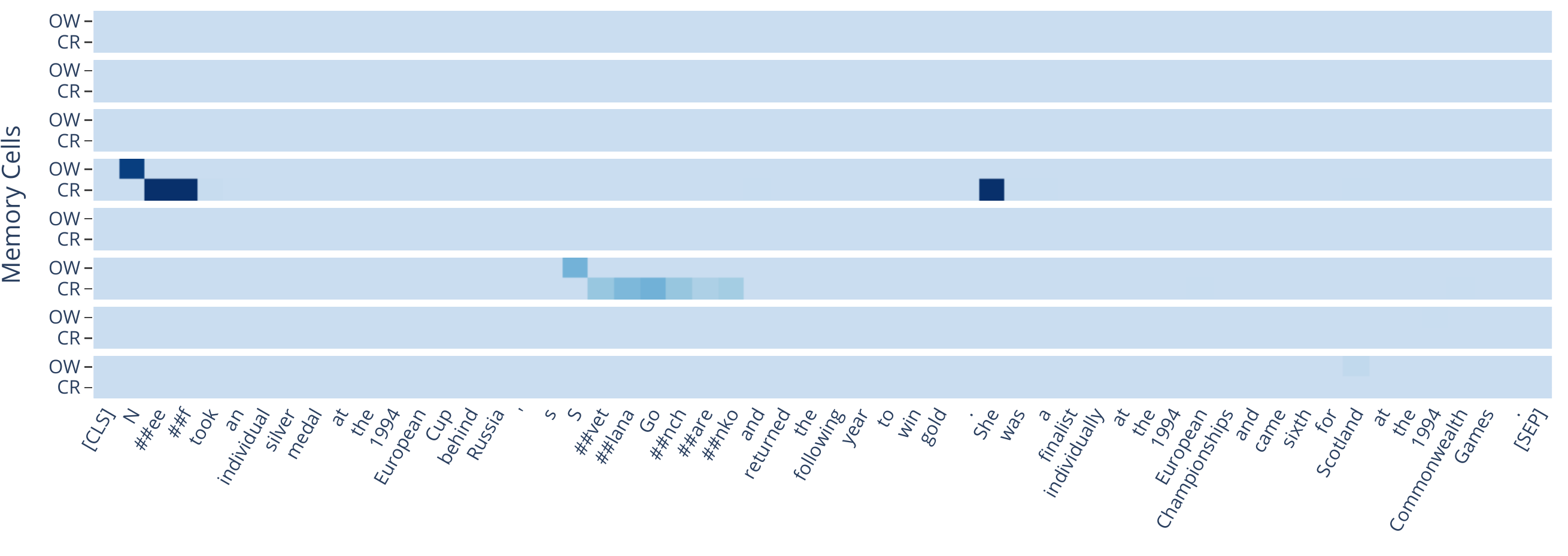}
    \caption{Memory log of \modelname with 8 memory cells. \modelname uses only 2 memory cells for the 2 unique people, namely Neef and Svetlana Goncharenko, and correctly resolves the pronoun.}
    \end{subfigure}
    \begin{subfigure}[t]{\textwidth}
    \includegraphics[width=\textwidth]{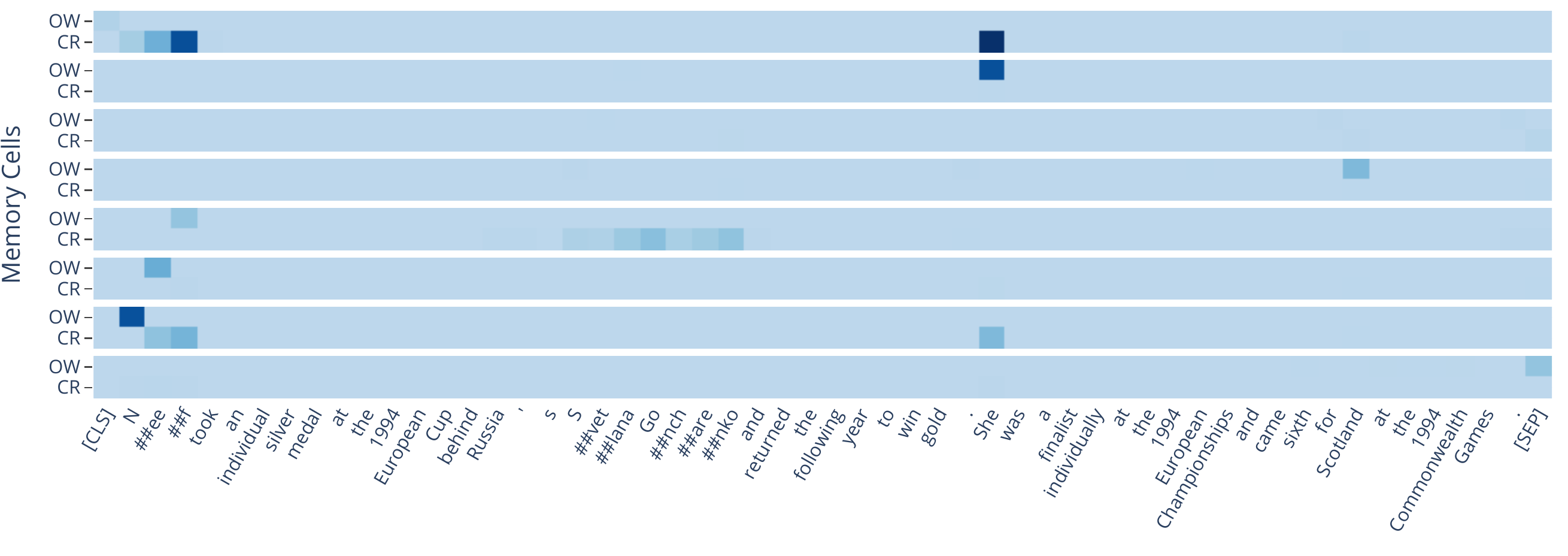}
    \caption{Memory log of the Referential Reader with 8-memory cells. The Referential Reader does successfully resolve the pronoun in the topmost memory cell but it ends up tracking Neef in as many as 4 memory cells.}
    \end{subfigure}
    \caption{
    Both the models only weakly detect ``Svetlana Goncharenko" which could be due to lack of span modeling.}
    \label{fig:petra_vs_ref_1}
\end{figure*}

\begin{figure*}[h]
    \begin{subfigure}[t]{\textwidth}
    \begin{mdframed}
        \hlent{Fripp}{1} has performed Soundscapes in several situations: * \hlent{Fripp}{1} has featured Soundscapes on various King Crimson albums. \hlent{He}{1} has also released pure Soundscape recordings as well: * On May 4, 2006, \hlent{\textcolor{red}{\it Steve Ball}}{2} invited \hlent{\textcolor{aqua}{\bf Robert Fripp}}{1} back to the Microsoft campus for a second full day of work on Windows Vista following up on \hlent{\textcolor{aqua}{\bf his}}{1} first visit in the Fall of 2005.
    \end{mdframed}
    \caption{GAP validation instance 17. The ground truth GAP annotation is indicated via colors.}
    \end{subfigure}
    \centering
    \begin{subfigure}[t]{\textwidth}
    \includegraphics[width=\textwidth]{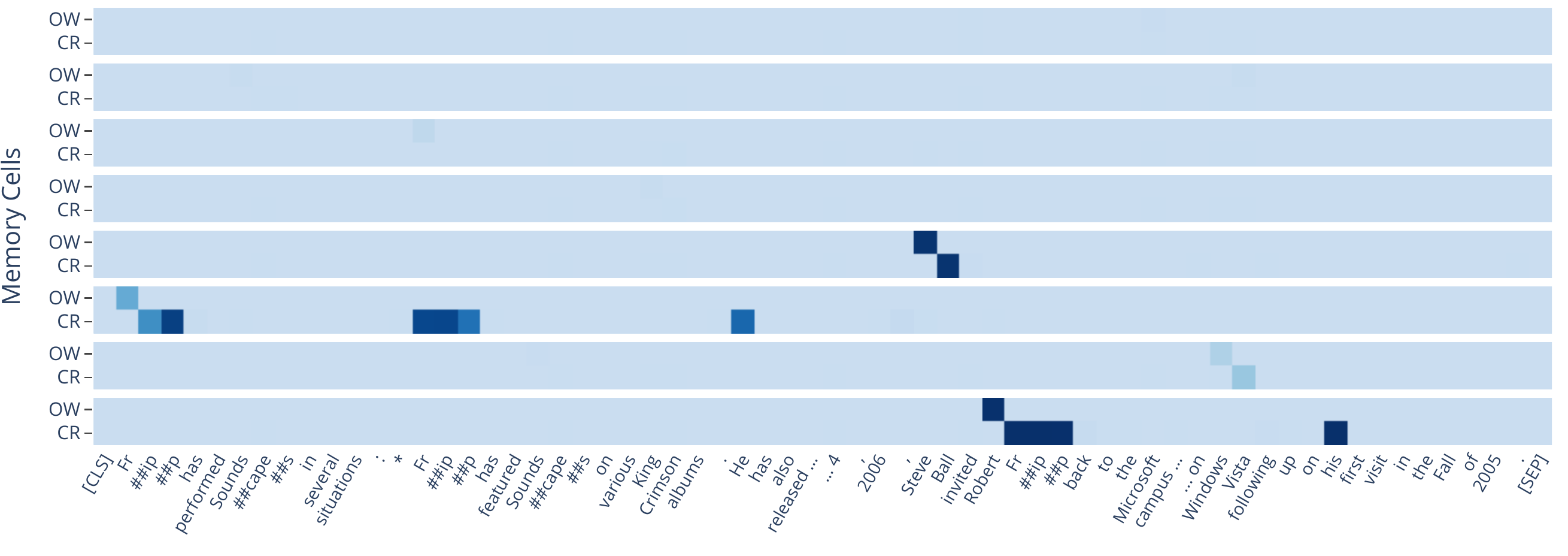}
    \caption{Memory log of \modelname with 8-memory cells. \modelname is pretty accurate at tracking Robert Fripp but it misses out on connecting ``Fripp" from the earlier part of the document to ``Robert Fripp". }
    \end{subfigure}
    \begin{subfigure}[t]{\textwidth}
    \includegraphics[width=\textwidth]{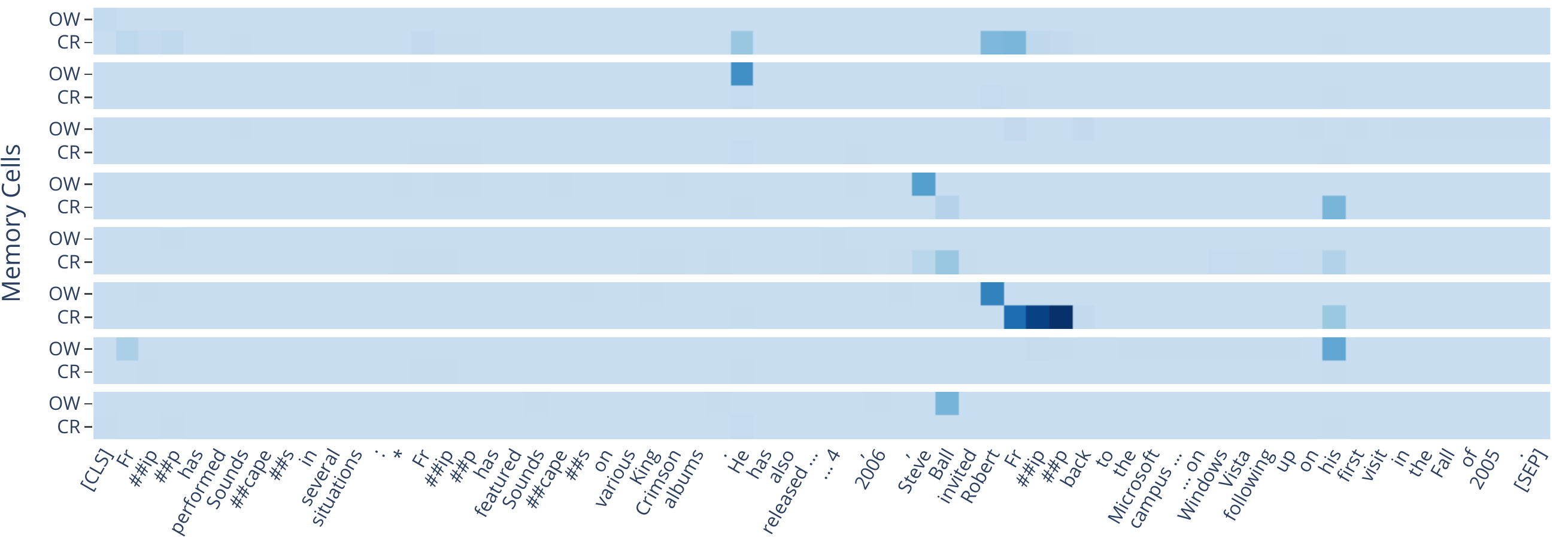}
    \caption{Memory log of the Referential Reader with 8-memory cells. The Referential Reader completely misses out on all the mentions in the first half of the document (which is not penalized in GAP evaluations where the relevant annotations are typically towards the end of the document). Apart from this, the model ends up tracking Robert Fripp in as many as 6 memory cells, and Steve Ball in 3 memory cells.}
    \end{subfigure}
    \caption{\modelname clearly performs better than the Referential Reader at people tracking for this instance. \modelname's output is more sparse, detects more relevant mentions, and is better at maintaining a 1-to-1 correspondence between memory cells and people.}
    \label{fig:petra_vs_ref_2}
\end{figure*}

Figure~\ref{fig:petra_vs_ref_1} and ~\ref{fig:petra_vs_ref_2} compare the memory logs of \modelname and the Referential Reader.

\end{document}